\pdfoutput=1

\documentclass[11pt]{article}

\usepackage[]{ACL2023}

\usepackage{times}
\usepackage{latexsym}

\usepackage[T1]{fontenc}

\usepackage[utf8]{inputenc}

\usepackage{microtype}

\usepackage{inconsolata}
\usepackage{amssymb}
\usepackage{amsmath} 
\usepackage{booktabs}
\usepackage{enumerate}
\usepackage{graphicx}
\usepackage{subfigure}
\usepackage{xspace}
\usepackage{float}
\usepackage{bbm}
\usepackage{bm}
\usepackage{multirow}
\usepackage{booktabs}
\usepackage{color}
\usepackage{framed}
\usepackage{stfloats}
\usepackage{iitem}
\usepackage{makecell}
\definecolor{shadecolor}{RGB}{180,180,180}
\usepackage{colortbl}
\usepackage{color, xcolor}

\newcommand{\paratitle}[1]{\vspace{1.5ex}\noindent\textbf{#1}}
\newcommand\Vector{\bm}
\newcommand\Matrix{\mathbf} 
\newcommand\Tensor{\mathcal}
\newcommand{\ie}{\emph{i.e.,}\xspace}

\newcommand{\eg}{\emph{e.g.,}\xspace}

\newcommand{\ignore}[1]{}

\usepackage{algorithm}
\newtheorem{theorem}{Theorem}[section]
\newtheorem{corollary}[theorem]{Corollary}
\usepackage{algpseudocode}
\usepackage{amsmath}
\usepackage{tikz}

\title{Scaling Pre-trained Language Models to Deeper \\ via Parameter-efficient Architecture}

\author{
	Peiyu Liu$^{1,3}$\thanks{$\ $ Authors contributed equally.},
	Ze-Feng Gao$^{1*}$,
        Yushuo Chen$^{1,3}$,
	Wayne Xin Zhao$^{1,3}$\thanks{$\ $ Corresponding author.},\and
	\textbf{Ji-Rong Wen}$^{1,2,3}$
	\\
	$^1$Gaoling School of Artificial Intelligence, Renmin University of China\\
	$^2$ School of Information, Renmin University of China\\
	$^3$Beijing Key Laboratory of Big Data Management and Analysis Methods\\
	{\tt\{liupeiyustu,zfgao,jrwen\}@ruc.edu.cn, }\\ 
	{\tt batmanfly@gmail.com,chenyushuo1999@foxmail.com}
}

\begin{document}
\maketitle

\begin{abstract}
In this paper, we propose a highly parameter-efficient approach to scaling pre-trained language models~(PLMs) to a deeper model depth. 
Unlike prior work that shares all parameters or uses extra blocks, we design a more capable parameter-sharing architecture based on  matrix product operator~(MPO). 
MPO decomposition can reorganize and factorize the information of a parameter matrix into two parts: the major part that contains the major information (\emph{central tensor}) and the supplementary part that only has a small proportion of parameters (\emph{auxiliary tensors}). Based on such a decomposition,  our architecture shares the central tensor 
across all layers for reducing the model size and meanwhile keeps layer-specific auxiliary tensors (also using adapters) for enhancing  the adaptation flexibility.   To improve the model training,  
we further propose a stable initialization algorithm tailored for the MPO-based architecture.
Extensive experiments have demonstrated the effectiveness of our proposed model in reducing the model size and achieving highly competitive performance. 

\end{abstract}

\section{Introduction}
Recently,  pre-trained language models~(PLMs) have achieved huge success in a variety of NLP tasks by exploring ever \emph{larger} model architecture~\cite{raffel2020exploring, radford2019language}.
It has been shown that there potentially exists a scaling law between the model size and model capacity for PLMs~\cite{kaplan2020scaling}, attracting many efforts to enhance the performance by scaling model size~\citep{chowdhery2022palm, wang@2022deepnet}. 

\begin{figure}[t]
    \includegraphics[width=0.5\textwidth]{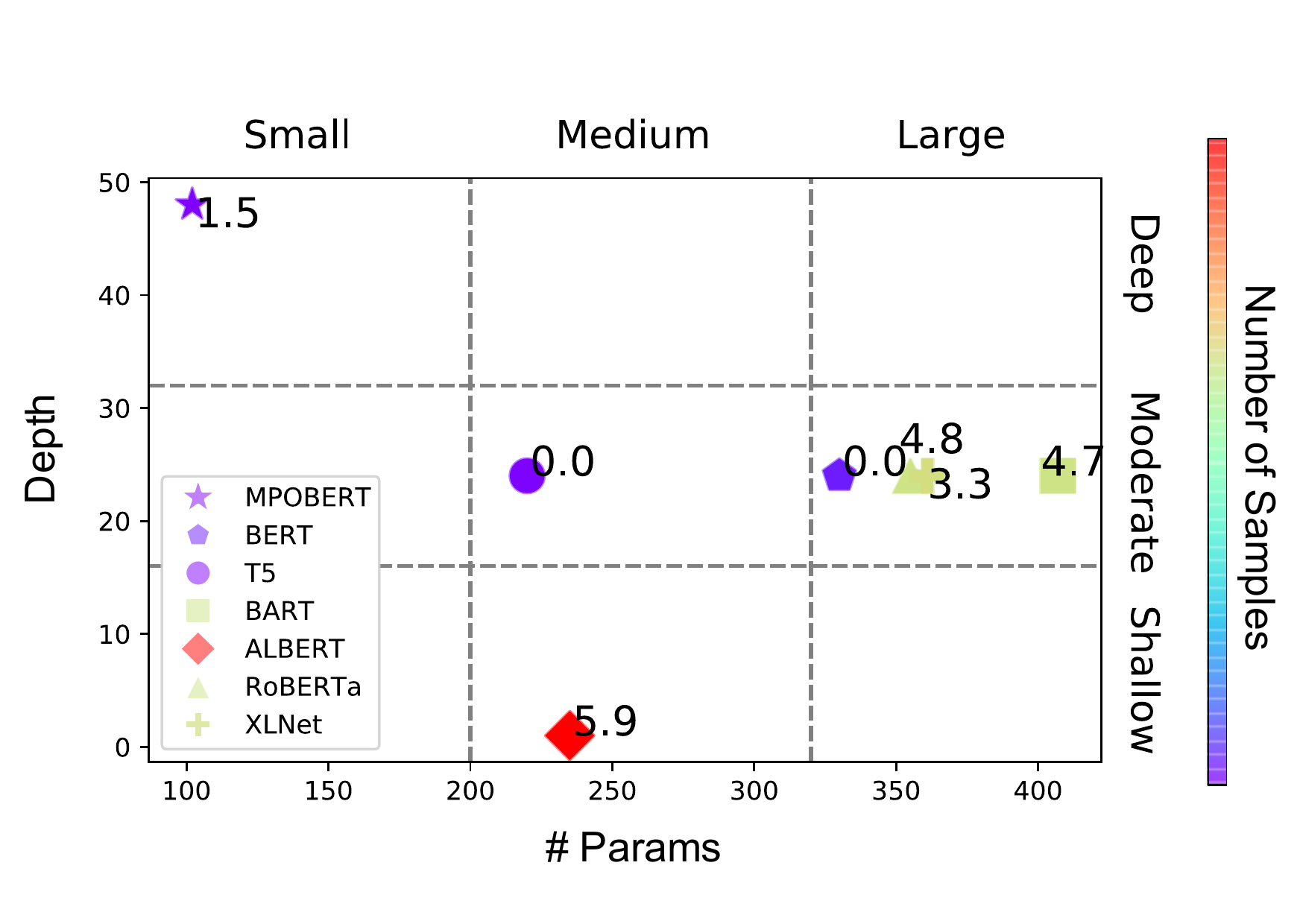}
    \caption{A comparison of our model and representative PLMs in the three dimensions of \emph{model size}, \emph{model depth}, and \emph{performance}~(measured on GLUE score). }
    \label{fig:plms_comparison}
\end{figure}
As a straightforward approach, we can directly increase the layer number of Transformer networks  for improving the model capacity~\cite{wang@2022deepnet,huang2020initialize}. 
While, a very deep architecture typically corresponds to a significantly large model size, leading to high costs in both computation and storage~\cite{gong@2019efficient_stack}.  
And, it is difficult to deploy deep networks in resource-limited settings, though it usually has a stronger model capacity.  
Therefore, there is an urgent need for developing a parameter-efficient way for scaling the model depth. 

To reduce the parameters in deep networks,  
weight sharing has proven to be very useful to design lightweight Transformer architectures~\citep{zhang2022minivit,lan2019albert}. As a representative work by across-layer parameter sharing, ALBERT~\citep{lan2019albert}  keeps only ten percent of the whole parameters of BERT  while maintaining comparable performance.
Although the idea of parameter sharing is simple yet (to some extent) effective, 
 it has been found that identical weights across different layers are the main cause of performance degradation~\cite{zhang2022minivit}. 
To address this issue, extra blocks are designed to elevate parameter diversity in each layer~\citep{nouriboriji@2022minialbert}. While, they still use the rigid architecture of 
shared layer weights,  having a limited model capacity. 
Besides, it is difficult to optimize very deep models, especially when shared components are involved. Although recent studies~\citep{wang@2022deepnet,huang2020initialize} propose  improved  initialization methods, they do not consider the case with parameter sharing, thus likely leading to a suboptimal performance on a parameter-sharing architecture. 
To address these challenges, in this paper, we propose a highly parameter-efficient approach to scaling PLMs to a deeper model architecture. 
As the core contribution, we propose a \emph{matrix product operator}~(MPO) based parameter-sharing architecture for deep Transformer networks.  
Via MPO decomposition,  a parameter matrix can be decomposed into  \emph{central tensors}~(containing the major information) and \emph{auxiliary tensors}~(containing the supplementary information). 
Our approach shares the central tensors of the parameter matrices across all layers for reducing the model size, and meanwhile keeps layer-specific auxiliary tensors for enhancing the adaptation flexibility.  
In order to train such a deep architecture, we propose  an MPO-based initialization method by utilizing the MPO decomposition results of ALBERT. Further, for the auxiliary tensors of  higher layers (more than 24 layers in ALBERT), we propose to set the parameters with scaling coefficients derived from theoretical analysis. We theoretically show it can address the training instability regardless of the model depth. 

Our work  provides a novel parameter-sharing way for scaling model depth, which can be generally applied to  various Transformer based  models. 
We conduct extensive  experiments to evaluate the performance of the proposed MPOBERT model on the GLUE benchmark in comparison to PLMs with varied model sizes~(tiny, small and large). 
Experiments  results have demonstrated  the effectiveness of the proposed model in reducing the model size and achieving competitive performance. With fewer parameters than BERT$_{\rm{BASE}}$, we scale the model depth by a factor of 4x and achieve 0.1 points higher than  BERT$_{\rm{LARGE}}$ for GLUE score. 



\section{Related Work}



\paratitle{Matrix Product Operators}.
Matrix product operators~(\emph{a.k.a.} tensor-train operators~\citep{oseledets2011tensor}) were proposed for a more effective representation of the linear structure of neural networks~\citep{gao2020compressing}, which was then used to compress deep neural networks~\citep{novikov2015tensorizing}, convolutional neural networks~\citep{garipov2016ultimate, yu2017long}, and LSTM~\citep{gao2020compressinglstm,sun2020model}.
Based on MPO decomposition, recent studies designed lightweight fine-tuning and compression methods for PLMs~\citep{liu2021enabling}, and developed  
parameter-efficient MoE architecture~\citep{gao2022parameter}.
Different from these works, our work aims to develop a very deep PLM  with lightweight architecture and stable training. 

\paratitle{Parameter-Efficient PLMs}.
Existing efforts to reduce the parameters of PLMs can be broadly categorized into three major lines: knowledge distillation, model pruning, and parameter sharing. For knowledge distillation-based methods~\cite{sanh2019distilbert,sun2020mobilebert,sun2020mobilebert,liu2020fastbert}, PLMs were distilled into student networks with much fewer parameters. For pruning-based methods, they tried to remove less important components~\cite{michel2019prunehead,wang2020prunestructure} or very small weights~\cite{chen2020lotterybert}. 
Moreover, the parameter-sharing method was further proposed by sharing all parameters~\cite{lan2019albert} or incorporating specific auxiliary components~\cite{reid-etal-2021-subformer-exploring,nouriboriji@2022minialbert}. Different from these works, we design an MPO-based architecture that can reduce the model size  and enable adaptation flexibility, by decomposing the original matrix. 


\paratitle{Optimization for Deep Models}.
Although it is simple to increase the number of layers for scaling model size, it is difficult to optimize very deep networks due to the training instability issue. Several studies have proposed different strategies to overcome this difficulty for training deep Transformer networks, including Fixup~\cite{zhang2019fixup} by  properly rescaling  standard initialization, T-Fixup~\cite{huang2020initialize} by proposing a weight initialization scheme, and DeepNorm~\cite{wang@2022deepnet} by introducing new normalization function. 
As a comparison, we study how to optimize the deep MPO-based architecture 
with the parameter sharing strategy, and explore the use of well-trained PLMs for initialization, which has a different focus from existing work. 

\ignore{The general way to obtain a deep model that is to increase the number of layers.
However, the optimization instability issues make it hard to train directly.
Several approaches have been proposed to address the issues of optimization instability. 
\citet{zhang2019fixup} proposed the Fixup initialization method to properly rescale standard initialization. 
Moreover, T-Fixup~\citep{huang2020initialize} presented a weight initialization scheme for the Transformer that eliminates the need for both layer normalization and warmup by addressing instability in the Adam optimizer. 
Another approach was proposed in~\citet{wang@2022deepnet} where they introduced a new normalization function~(DEEPNORM) to modify the residual connection in the Transformer and a theoretically derived initialization method that allows for scaling Transformers up to 1,000 layers. 
These methods prove to be effective, however, the high computational costs associated with pre-training from scratch can limit their application in resource-limited settings. 
In comparison, we propose methods to solve the optimization problem while also allowing the training process to save computational and storage costs.
}

\section{Method}
In this section, we describe the proposed \emph{MPOBERT} approach for building deep PLMs via a highly parameter-efficient architecture.
Our approach follows the classic \emph{weight sharing} paradigm, while introducing a  principled mechanism for sharing informative parameters across layers and also enabling  layer-specific  weight adaptation. 

\subsection{Overview of Our Approach}
\label{sec-framework}
Although weight sharing has been widely explored for building  compact PLMs~\cite{lan2019albert}, existing studies either  share  all the parameters across  layers~\cite{lan2019albert} or incorporate additional blocks to facilitate  the sharing~\cite{zhang2022minivit,nouriboriji@2022minialbert}. They  either have limited model capacity with a rigid architecture  or require additional efforts for maintenance.   

Considering the above issues, we motivate our approach in two aspects.  Firstly, only informative parameters should be shared across layers, instead of all the parameters.  Second, it should not affect the capacity to capture layer-specific  variations. 
To achieve this, we utilize the MPO decomposition from multi-body physics~\cite{gao2020compressing} to develop a parameter-efficient architecture by sharing informative components  across layers and keeping layer-specific supplementary components  (Section~\ref{subsec-crosslayer}). 
As another potential issue, it  is difficult to optimize  deep PLMs due to unstable training~\cite{wang@2022deepnet}, especially when weight sharing~\cite{lan2019albert} is involved. 
We further propose a simple yet effective method to stabilize the training of MPOBERT (Section~\ref{sec-efficient-training}).  Next, we introduce the technical details of our approach.

\subsection{MPO-based Transformer Layer}
\label{subsec-crosslayer}
In this section, we first introduce the MPO decomposition and introduce how to utilize it for building parameter-efficient deep PLMs.  

\subsubsection{MPO Decomposition}


Given a weight matrix $\Matrix{W}\in \mathbb{R}^{I\times J}$,   MPO decomposition~\cite{gao2020compressing} can  decompose a matrix into a product of $n$ tensors by reshaping the two dimension sizes $I$ and $J$:
\begin{equation}
    \small
   \Matrix{W}_{i_1,\dots , i_n, j_1, \dots, j_n} = \Tensor{T}^{(1)}[i_1, j_1]\cdots \Tensor{T}^{(n)}[i_n, j_n],
\label{eq:mpo-decom}
\end{equation}
where we have $I=\prod_{k=1}^{n} i_k$, $J=\quad \prod_{k=1}^{n}j_k$, and  $\Tensor{T}^{(k)}[i_k, j_k]$ is a 4-dimensional tensor with size $d_{k-1}\times i_k \times j_k \times d_k$ in which $d_k$ is a bond dimension linking $T^{(k)}$ and $T^{(k+1)}$ with $d_0=d_n=1$. For simplicity, we omit the bond dimensions  in Eq.~\eqref{eq:mpo-decom}.
When $n$ is odd, the middle tensor contains the most parameters (with the largest bond dimensions), while the parameter sizes of the rest decrease with the increasing distance to the middle tensor. 
Following \citep{liu2021enabling}, we further simplify the decomposition results of a matrix as a central tensor  $\mathcal{C}$ (the middle tensor) and auxiliary tensors $\{ \mathcal{A}_i \}_{i=1}^{n-1}$ (the rest tensor).

As a major merit, such a decomposition
can effectively reorganize and aggregate the information of the matrix~\cite{gao2020compressing}:   central tensor  $\mathcal{C}$ can encode the essential information of the original matrix, while  auxiliary tensors $\{ \mathcal{A}_i \}_{i=1}^{n-1}$ serve as its complement to exactly reconstruct the matrix.  

\subsubsection{MPO-based Scaling to Deep Models }
\label{subsec-mpobased_scaling}
\begin{figure}[t]
    \centering
    \hspace{-0.33cm}\includegraphics[width=0.5\textwidth]{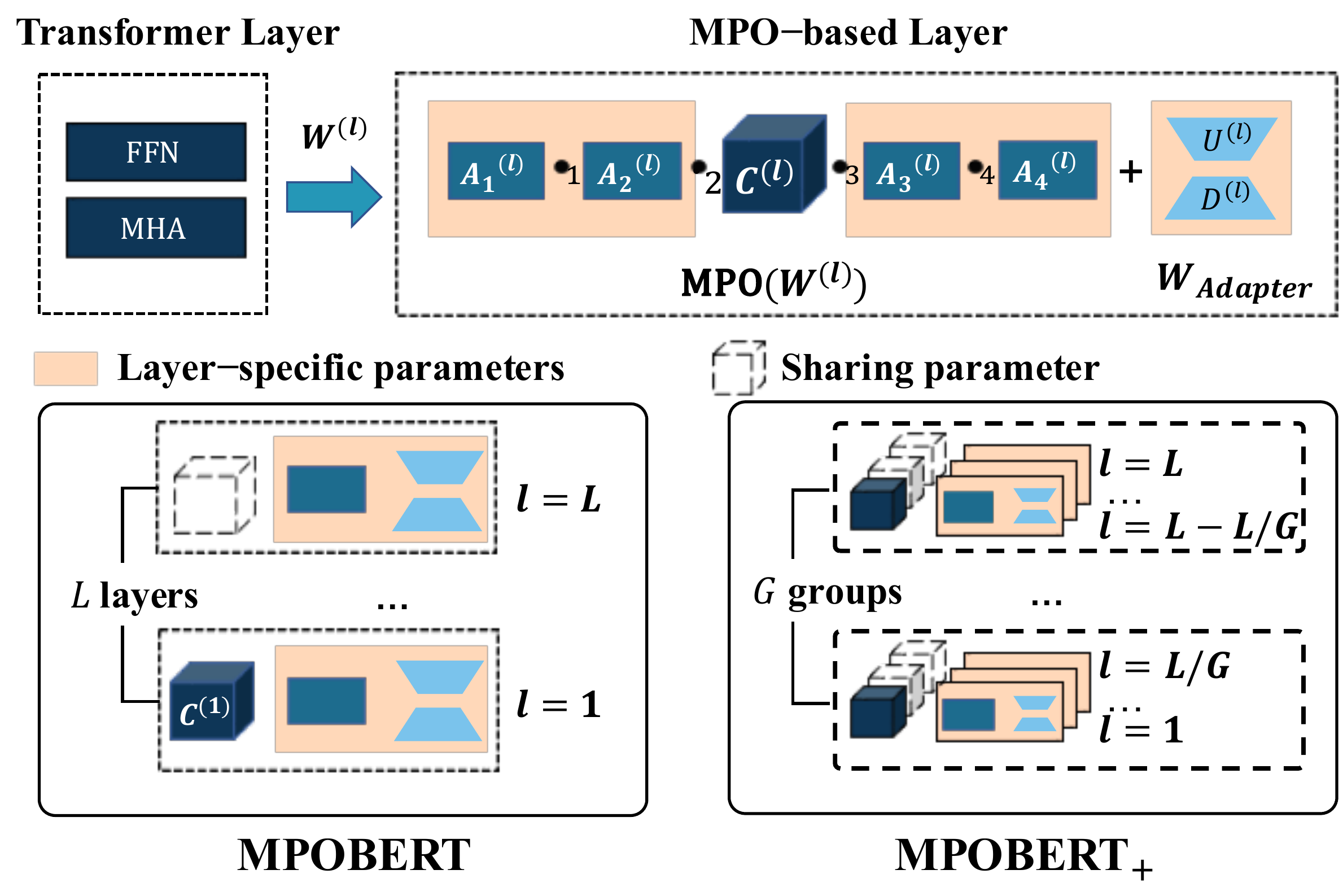}
    \caption{Overview architecture of MPOBERT and MPOBERT$_{+}$. We use blocks with dashed borderlines to represent shared central tensors. Central tensors are shared across all $L$ Layers in MPOBERT and within groups in MPOBERT$_{+}$.}
    \label{fig:main}
\end{figure}

Based on MPO decomposition, the essence of our scaling method is to share the central tensor across layers (\emph{capturing the essential information}) and keep layer-specific auxiliary tensors (\emph{modeling layer-specific  variations}). Fig.~\ref{fig:main} shows the overview architecture of the proposed MPOBERT. 


\paratitle{Cross-layer Parameter Sharing}. To introduce our architecture, we  consider a simplified structure of $L$ layers, each consisting of a single matrix. With the five-order MPO decomposition (\ie $n=5$), we can obtain the decomposition results for a weight matrix ($\Matrix{W}^{(l)}$), denoted as $\{\Tensor{C}^{(l)}, \Tensor{A}_1^{(l)}, \Tensor{A}_2^{(l)}, \Tensor{A}_3^{(l)}, \Tensor{A}_4^{(l)}\}_{l=1}^{L}$, where   $\Tensor{C}^{(l)}$ and $\{\Tensor{A}_i^{(l)}\}_{i=1}^{4}$ are the central tensor and auxiliary tensors of the $l$-th layer. Our approach is to set a shared central tensor $\Tensor{C}$ across layers, which means that  $\Tensor{C}^{(l)}=\Tensor{C}$ ($\forall l=1\cdots L$). 
As shown in \citet{gao2020compressing}, the central tensor contains the major proportion of  parameters (more than 90\%), and thus our method can largely reduce the parameters when scaling a PLM to very deep architecture.  Note that this strategy 
can be easily applied to multiple matrices in a Transformer layer, and we omit the discussion for the multi-matrix extension.  Another extension is to share the central tensor by different grouping layers. 
We implement a layer-grouping MPOBERT, named \emph{MPOBERT$_{+}$}, which divides the layers into multiple parts and sets unique shared central tensors in each group.

\paratitle{Layer-specific Weight Adaptation}. Unlike ALBERT~\cite{lan2019albert}, our MPO-based architecture  enables layer-specific adaptation by keeping layer-specific auxiliary tensors ($\{\Tensor{A}_i^{(l)}\}_{i=1}^{4}$). 
These auxiliary tensors are decomposed from the original matrix, instead of extra blocks~\cite{zhang2022minivit}. They only contain a very small proportion of parameters, which does not significantly increase the model size. While, another merit of MPO decomposition is that these tensors are highly correlated via bond dimensions, and a small perturbation on an auxiliary tensor can reflect the whole matrix~\cite{liu2021enabling}.  If the downstream task requires more layer specificity, we can further  incorporate low-rank adapters~\cite{hu2021lora} for layer-specific adaptation. Specifically,  
we denote $\Matrix{W}^{(l)}_{Adapter}$ as the low-rank adapter for $\Matrix{W}^{(l)}$. 
In this way, $\Matrix{W}^{(l)}$ can be formulated as a set of tensors: $\{\Tensor{C}^{(l)},  \Tensor{A}_1^{(l)}, \Tensor{A}_2^{(l)},  \Tensor{A}_3^{(l)},  \Tensor{A}_4^{(l)}, \Matrix{W}^{(l)}_{Adapter}\}$. The parameter scale of  adapters,  $L \times r \times d_{total}$, is determined by the layer number $L$, the rank $r$, and the shape of  the original matrix ($d_{total}=d_{in}+d_{out}$ is the sum of the input and output dimensions of a Transformer Layer). Since we employ low-rank adapters, we can effectively control the number of  additional parameters from  adapters.

\subsection{Stable Training for MPOBERT}
\label{sec-efficient-training}

With the above MPO-based approach, we can scale a PLM to a deeper architecture  in a highly  parameter-efficient way. 
However, as shown in prior studies~\cite{lan2019albert,wang@2022deepnet}, it is difficult to optimize very deep PLMs, especially when shared components are involved. In this section, we introduce a simple yet stable training algorithm for MPOBERT and then discuss how it addresses the training instability issue.  

\subsubsection{MPO-based Network Initialization}
\label{sec-mpo-based-network-initialization}
Existing work has found that parameter initialization is important for training deep models~\cite{huang2020initialize,zhang2019fixup,wang@2022deepnet}, which can help  alleviate the training instability. 
To better optimize the scaling MPOBERT, we propose a specially  designed  initialization method  based on the above MPO-based architecture.

\paratitle{Initialization with MPO Decomposition}. Since MPOBERT shares global components (\ie the central tensor) across all layers, our idea is to employ  existing well-trained PLMs based on weight sharing for improving parameter initialization.  Here, we use the released 24-layer ALBERT with all the parameters shared across layers. The key idea is to perform MPO decomposition on the parameter matrices of ALBERT, and obtain the corresponding central and auxiliary tensors. Next, we discuss the initialization of MPOBERT in two aspects. 
 For \emph{central tensors}, we directly initialize them  (each for each matrix) by the derived central tensors from the MPO decomposition results of ALBERT. Since they are globally shared, one single copy is only needed for initialization regardless of the layer depth. Similarly,  for  \emph{auxiliary tensors}, we can directly copy the auxiliary tensors from the  MPO decomposition results of ALBERT.
 
\paratitle{Scaling the Initialization}. A potential issue is that ALBERT only provides a 24-layer architecture, and such a strategy no longer supports the  initialization for an architecture of more than 24 layers (without corresponding auxiliary tensors). As our solution, we borrow the idea in \citet{wang@2022deepnet} that  avoids the exploding update by incorporating an additional scaling coefficient and multiplying  the randomly initialized values for the auxiliary tensors (those in higher than 24 layers) with a coefficient of $(2L)^{-\frac{1}{4}}$, where $L$ is the layer number. Next, we present a theoretical analysis  of  training stability.

\subsubsection{Theoretical Analysis}

To understand the issue of training instability from a theoretical perspective, we consider a Transformer-based model $F(\Vector{x},\Matrix{W})$ with $\Vector{x}$ and $\Matrix{W}$ as input and parameters, and consider one-step update $\bigtriangleup F$\footnote{$\bigtriangleup F \overset{\bigtriangleup}{=}F(\Vector{x},\Matrix{W}-\eta\frac{\partial}{\partial \Matrix{W}}\mathcal{L}(F(\Vector{x})-y))-F(\Vector{x};\Matrix{W}).$}. 
\ignore{
\begin{equation}
\small
\bigtriangleup F \overset{\bigtriangleup}{=}F(\Vector{x},\Matrix{W}-\eta\frac{\partial}{\partial \Matrix{W}}\mathcal{L}(F(\Vector{x})-y))-F(\Vector{x};\Matrix{W}).
\end{equation} 
}
According to \citet{wang@2022deepnet}, a large model update ($\bigtriangleup F$) at the beginning of training is likely to cause the training instability of deep Transformer models. 
To mitigate the exploding update problem,  the update should be bounded by a constant, \ie $\left\| \bigtriangleup F\right\|=\mathcal{O}(1)$.
Next, we study how the $\bigtriangleup F$ is bounded with the MPOBERT. 

\ignore{
\textcolor{blue}{We first express the model updates in terms of $\bigtriangleup F$. That is, }given a learning rate $\eta$, the update to the model can be written as 
$\bigtriangleup F \overset{\bigtriangleup}{=}F(\Vector{x},\Matrix{W}-\eta\frac{\partial}{\partial \Matrix{W}}\mathcal{L}(F(\Vector{x})-y))-F(\Vector{x};\Matrix{W})$. 
\textcolor{blue}{~\citet{wang@2022deepnet} demonstrates the instability starts from the large model update, \ie $\left\| \bigtriangleup F\right\|$, at the beginning of training. In order to mitigate the exploding update problem, we need to make the update bounded by a constant, \ie $\left\| \bigtriangleup F\right\|=\mathcal{O}(1)$.}
}

\paratitle{MPO-based Update Bound}. Without loss of generality, we consider a simple case of low-order MPO decomposition: $n=3$ in Eq.~\eqref{eq:mpo-decom}.  Following the derivation method in \citet{wang@2022deepnet}, we simplify the matrices $\Matrix{W}$, $\Tensor{A}_1$,  $\Tensor{C}$ and $\Tensor{A}_2$ to scalars $w$,$u$,$c$,$v$, which means the parameter $w_l$ at the $l$-th layer   can be decomposed as $w_l=u_l\cdot c_l\cdot v_l$. Based on these notations, we consider $L$-layer Transformer-based model $F(x,w)(w=\{w_1, w_2, ...,w_L\})$, where each sub-layer is normalized with Post-LN: $x_{l+1}=LN(x_l+G_l(x_l,w_l))$. Then we can prove $\left\| \bigtriangleup F\right\|$ satisfies (see Theorem~\ref{app-thm1} in the Appendix):
       \begin{align}
    \left\| \bigtriangleup F\right\|\leq
    &\sum_{l=1}^{L}(c_1v_l \left\|u_{l}^*-u_{l}\right\|+ c_1u_l \left\|v_{l}^*-v_{l}\right\| \nonumber\\
    &+ v_lu_l \left\|c_1^*-c_1\right\|),
    \label{eq-theorem1}
    \end{align}
\ignore{
\begin{theorem}
    Given an $L$-layer Transformer-based model $F(x,w)(w=\{w_1, w_2, ...,w_L\})$, where $w_l$ denotes the parameters in $l$-th layer and each sub-layer is normalized with Post-LN: $x_{l+1}=LN(x_l+G_l(x_l,w_l))$. In MPOBERT, $w_l$ is decomposed by MPO to local tensors: $w_l=u_l\cdot c_l\cdot v_l$, and we share $\{c_i\}_{i=1}^{L}$ across $L$ layers: $c_l=c_1, l=1,2,\cdots,L$. Then $\left\| \bigtriangleup F\right\|$ satisfies:
\label{theorem-1}
\end{theorem}
}
The above equation bounds the model update in terms of the central and auxiliary tensors. 
Since central tensors ($c_l$) can be initialized using the pre-trained weights, we can further simplify the above bound by reducing them. With some derivations (See Corollary~\ref{app-thm2} in the Appendix), we can obtain $(v_i^2+u_i^2)(u_Lv_L)=\mathcal{O}(\frac{1}{L})$ in order to guarantee that $\left\| \bigtriangleup F\right\|=\mathcal{O}(1)$.   For simplicity, we set $u_i=v_i=(2L)^{-\frac{1}{4}}$ to  bound the magnitude of each update  independent of layer number $L$. In the implementation, we first adopt the Xavier method for initialization, and then scale the parameter values with the coefficient of $(2L)^{-\frac{1}{4}}$.

\paratitle{Comparison}. Previous research has shown that using designed values for random initialization can improve the training of deep models~\citep{huang2020initialize,zhang2019fixup,wang@2022deepnet}. These methods  aim to 
improve the initialization of general Transformer architectures  for training from scratch. 
As a comparison, we explore the use of pre-trained weights  and  employ the MPO decomposition results for initialization.  
In particular,~\citet{gong@2019efficient_stack} have demonstrated the effectiveness of stacking pre-trained shallow layers for deep models in  accelerating convergence, also showing performance  superiority  of pre-trained weights over random initialization.


\subsubsection{Training and Acceleration}
\ignore{In general, our approach can be easily adapted to various PLMs for scaling along the model depth and efficient training~(see Algorithm~\ref{alg-overall-process}). 
We implemented various strategies to reduce the pre-training cost of MPOBERT during the process of realization. The results are summarized in Table~\ref{tab-accelerating}.
}

To instantiate our approach, we pre-train a 48-layer BERT model (\ie MPOBERT$_{48}$). For a fair comparison with BERT$_{\rm{BASE}}$ and BERT$_{\rm{LARGE}}$, we adopt the same pre-training corpus~(BOOKCORPUS~\citep{zhu2015aligning} and English Wikipedia~\citep{devlin2018bert}) and pre-training tasks~(masked language modeling, and sentence-order prediction).  
We first perform MPO decomposition on the weights of ALBERT and employ the initialization algorithm in Section~\ref{sec-mpo-based-network-initialization} to set the parameter weights. During the training, we need to keep an updated copy of central tensors and auxiliary tensors: we optimize them according to the pre-training tasks in an end-to-end way and combine  them to derive the original parameter matrix for forward computation (taking a relatively small cost of parallel matrix multiplication).

Typically, the speed of the pre-training process is affected by three major factors: arithmetic bandwidth, memory bandwidth, or latency. We further utilize a series of efficiency optimization ways to accelerate the pre-training, such as {mixed precision training with FP16} (reducing memory and arithmetic bandwidth) and {fused implementation of activation and normalization} (reducing latency). Finally, we can train the 48-layer MPOBERT at a time cost of 3.8 days (compared with a non-optimized cost of 12.5 days) on our server configuration (8 NVIDIA V100 GPU cards and 32GB memory). More training details are can be found in the experimental setup Section~\ref{sec-experimental-setup} and  Appendix~\ref{add-trianing-detail}~(Table~\ref{tab-strongest_variants} and Algorithm~\ref{alg-overall-process}).

\section{Experiments}
In this section, we first set up the experiments and then evaluate the efficiency of MPOBERT on a variety of tasks with different model settings.

\subsection{Experimental Setup}
\label{sec-experimental-setup}
\paratitle{Pre-training Setup}.
For the architecture, we denote the number of layers as $L$, the hidden size as $H$, and the number of self-attention heads as $A$. We report results on four model sizes: \textbf{MPOBERT$_{\textbf{12}}$}~($L$=12, $H$=768, $A$=12), \textbf{MPOBERT$_{\textbf{24}}$} ($L$=24, $H$=1024, $A$=16), \textbf{MPOBERT$_{\textbf{48}}$}~($L$=48, $H$=1024, $A$=16) and \textbf{MPOBERT$_{\textbf{48+}}$} that 
implement cross-layer parameter sharing in three distinct groups as discussed in subsection~\ref{subsec-mpobased_scaling}.
We pre-train all of the models with a batch size of 4096 for 10$k$ steps. Our code will be released after the review period.

\paratitle{Fine-tuning Datasets}.
To evaluate the performance of our model, we conduct experiments on the GLUE~\citep{wang2018glue} and SQuAD v1.1~\citep{rajpurkar2016squad} benchmarks.
Since fine-tuning is typically fast, we run an exhaustive parameter search and choose the model that performs best on the development set to make predictions on the test set. 
We include the details in the Appendix(see Appendix~\ref{add-detail_dataset} for the datasets and Appendix~\ref{add-detail_metric} for evaluation metrics)
\begin{table*}[ht]
\centering
\small
\begin{tabular}{l|rrrrrrrrr|rr}                                     
\toprule[1pt]
\small
\multirow{2}{*}{Experiments}    & \makebox[0.04\textwidth][c]{MRPC} & \makebox[0.04\textwidth][c]{SST-2} & \makebox[0.04\textwidth][c]{CoLA} & \makebox[0.04\textwidth][c]{RTE}   & \makebox[0.04\textwidth][c]{STS-B} & \makebox[0.04\textwidth][c]{QQP} & \makebox[0.04\textwidth][c]{MNLI} & \makebox[0.04\textwidth][c]{QNLI} & \makebox[0.04\textwidth][c]{SQuAD} &  \makebox[0.04\textwidth][c]{Avg.} &\makebox[0.04\textwidth][c]{\#To~(M)}  \\ 
                                & F1   & Acc.  & Mcc.   & Acc.   & Spear. & F1/Acc. & Acc. & Acc. & F1\\ \midrule
\rowcolor{gray!10}\multicolumn{12}{c}{\it \textbf{Development set}}\\
\rowcolor{gray!10}\multicolumn{12}{l}{\textbf{Tiny Models}~(\rm{\#To < 50M)}}\\
ALBERT$_{12}$                      & 89.0           & 90.6              & 53.4             & 71.1             & 88.2             & -/89.1             & 84.5             & 89.4             & 89.3          & 82.7             & 11 \\
ALBERT$_{24}$                      & 84.6           & \underline{93.6}  & 52.5             & \textbf{79.8}    & 90.1             & -/88.1             & 85.0             & \underline{91.7} & 90.6          & \underline{84.0} & 18 \\ 
\textcolor{purple}{MPOBERT$_{12}$} & \underline{90.3} & 92.3            & \underline{55.2} & 71.8             & \underline{90.5} & \underline{-/90.1} & \underline{84.7} & 91.2             & 90.1          & \underline{84.0} & 20 \\ 
\textcolor{purple}{MPOBERT$_{24}$} & \textbf{90.3}  & \textbf{94.4}     & \textbf{58.1}    & \underline{75.5} & \textbf{91.1}    & \textbf{-/90.2}    & \textbf{87.0}    & \textbf{92.6}    & \textbf{92.3} & \textbf{85.7}    & 46 \\\midrule
\rowcolor{gray!10}\multicolumn{12}{l}{\textbf{Small Models}~(\rm{50M < \#To < 100M)}}\\
T5$_{12}$                           & \underline{89.2}  & 94.7  & \underline{53.5}  & \underline{71.7} & \underline{91.2}   & \textbf{-/91.1}    & \textbf{87.8}    & \textbf{93.8}     & \underline{90.0}    & \underline{84.8} & 60\\ 
\textcolor{purple}{MPOBERT$_{48}$}  & \textbf{90.8}     & 94.7	& \textbf{58.3}     & \textbf{77.3}	   & \textbf{91.4}      & \underline{-/89.5} & \underline{86.3} & \underline{92.0}  & \textbf{92.3}       & \textbf{85.8}    & 75\\ \midrule
\rowcolor{gray!10}\multicolumn{12}{l}{\textbf{Base Models}~(\rm{\#To > 100M)}}\\
BERT$_{12}$                         & 90.7              & 91.7             & 48.9             & 71.4             & \underline{91.0} & \underline{-/90.8} & 83.7             & 89.3             & 88.5             & 82.9             & 110  \\
XLNet$_{12}$                        & 85.3              & \underline{94.4} & 49.3             & 63.9             & 85.6             & -/90.7             & \textbf{90.9}    & 91.8             & 90.2             & 82.5             & 117 \\
RoBERTa$_{12}$                      & \textbf{91.9}     & 92.2             & \textbf{59.4}    & 72.2             & 89.4             & \textbf{-/91.2}    & \underline{88.0} & \textbf{92.7}    & 91.2             & \underline{85.4} & 125\\
BART$_{12}$                         & \underline{91.4}  & 93.8             & 56.3             & \underline{79.1} & 89.9             & \underline{-/90.8} & 86.4             & \underline{92.4} & \textbf{91.6}    & 82.8             & 140 \\
\textcolor{purple}{MPOBERT$_{48+}$} & 89.7              & \textbf{94.4}    & \underline{57.4} & \textbf{79.8}    & \textbf{91.1}    & -/89.3             & 87.1             & \underline{92.4} & \underline{91.4} & \textbf{86.0}    & 102\\ \midrule\midrule
\rowcolor{gray!10}\multicolumn{12}{c}{\it \textbf{Test set}}\\
\rowcolor{gray!10}\multicolumn{12}{l}{\textbf{Tiny Models}~(\rm{\#To < 50M)}}\\
ALBERT$_{12}$                        & 89.2             & 93.2             & \underline{53.6} & 70.2            & \underline{87.3}  & 70.3/-            & 84.6              & \underline{92.5}   & 89.3             & 81.1              & 11\\
ALBERT$_{24}$                        & 88.7             & \underline{94.0} & 51.7	          & \textbf{73.7}	& 86.9              & 69.1/-	        & 84.9	            & 91.8	             & \underline{90.6}	& \underline{81.2}              & 18 \\
MobileBERT$_{24}$$\blacklozenge$     & 88.8             & 92.6             & 51.1             & 70.4            & 84.8              & \underline{70.5/-}& 83.3              & 91.6               & 90.3             & 80.4              & 25\\
\textcolor{purple}{MPOBERT$_{12}$}   & \textbf{89.2}    & 91.9             & 52.7             & 70.6            & 87.1              & 69.6/-            & \underline{85.0}  & 91.0               & 90.1             & 80.8              & 20     \\  
\textcolor{purple}{MPOBERT$_{24}$}   & \underline{89.0}	& \textbf{94.5}    & \textbf{55.5}	  & \underline{73.4}& \textbf{88.2}     & \textbf{71.0/-}	& \textbf{86.3}     & \textbf{93.0}      & \textbf{92.3}    & \textbf{82.6}     & 46   \\\midrule
\rowcolor{gray!10}\multicolumn{12}{l}{\textbf{Small Models}~(\rm{50M < \#To < 100M)}}\\
T5$_{12}$                           & \underline{89.7}  & 91.8              & 41.0             & 69.9               & 85.6              & 70.0/-            & 82.4              & 90.3              & 90.0              & 78.7              & 60\\ 
TinyBERT$_{6}$$\clubsuit$           & 87.3              & \underline{93.1}  & \underline{51.1} & \underline{70.0}   & \underline{83.7}  & \textbf{71.6/-}   & \underline{84.6}  & \underline{90.4}  & \underline{87.5}  & \underline{79.9}  & 67\\
DistilBERT$_{6}$$\clubsuit$         & 86.9              & 92.5              & 49.0             & 58.4               & 81.3              & 70.1/-            & 82.6              & 88.9              & 86.2              & 77.3              & 67\\
\textcolor{purple}{MPOBERT$_{48}$}  & \textbf{90.0}	    & \textbf{94.0}	    & \textbf{55.0}    & \textbf{74.0}	    & \textbf{88.7}     & \underline{71.0/-}& \textbf{86.5}     & \textbf{91.8}     & \textbf{92.3}     & \textbf{82.6}     & 75\\ \midrule
\rowcolor{gray!10}\multicolumn{12}{l}{\textbf{Base Models}~(\rm{\#To > 100M)}}\\
BERT$_{12}$$\spadesuit$              & 88.9             & 93.5              & 52.1          & 66.4              & 85.8              & 71.2/-            & 84.6             & 90.5              & 88.5               & 79.1     & 110    \\
XLNet$_{12}$                         & 89.2             & \underline{94.3}  & 47.3          & 66.5              & 85.4              & \underline{71.9/-}& \underline{87.1} & 91.4              & 90.2               & 80.4      & 117   \\
RoBERTa$_{12}$                       & 89.9	            & 93.2	            & \textbf{57.9} & 69.9	            & \underline{88.3}	& \textbf{72.5/-}   & \textbf{87.7}	   & \underline{92.5}	   & 91.2	& \underline{82.6}      & 125\\
BART$_{12}$                          & 89.9             & 93.7              & 49.6          & \underline{72.6}  & 86.9              & 71.7/-            & 84.9             & 92.3              & \textbf{91.6}               & 81.5      & 140   \\
\textcolor{purple}{MPOBERT$_{48+}$}  & \textbf{89.9}    & \textbf{94.5}     & \underline{56.0}  & \textbf{74.5} & \textbf{88.4}     & 70.5/-            & 86.5             & \textbf{92.6}  & \underline{91.4}      & \textbf{82.7}  & 102\\ \bottomrule
\end{tabular}
\caption{Performance comparison of different models on natural language understanding tasks~(in percent). ``\# To~(M)'' denote the number~(in millions) of total parameters. 
We compare MPOBERT with PLMs~(\ie BERT and ALBERT) and Parameter-efficient Transformers~(\ie MobileBERT, TinyBERT and DistilBERT), respectively. The best and the second-best performance in each task are highlighted in bold and underlined.
$\blacklozenge$: Experimental results by~\citet{sun2020mobilebert}.
$\clubsuit$: Experimental results by~\citet{jiao2019tinybert}.
$\spadesuit$: Experimental results by~\citet{devlin2018bert}.}
\label{tab-main_results}
\end{table*}

\paratitle{Baseline Models}.
We compare our proposed MPOBERT to the existing competitive deep PLMs and parameter-efficient models. In order to make fair comparisons, we divide the models into three major categories based on their model sizes: 

$\bullet$~{Tiny Models~(\rm{\#To < 50M}).} ALBERT$_{12}$~\cite{lan2019albert} is the most representative PLM that achieves competitive results with only 11M.

$\bullet$~{Small models~(50M< \#To <100M).}
We consider PLMs~(T5$_{12}$) and compressed models~(MobileBERT~\citep{sun2020mobilebert}, DistilBERT~\citep{sanh2019distilbert} and TinyBERT~\citep{jiao2019tinybert}).

$\bullet$~{Base models~(\#To > 100M).} We compare with BERT$_{12}$, XLNet$_{12}$, RoBERTa$_{12}$ and BART$_{12}$ for this category. 
Note that we only include the base variants that have similar model sizes in order to make a fair comparison. 

More details about the baseline models are described in Appendix~\ref{add-detail_baseline}. 

\subsection{Main Results}
\paratitle{Fully-supervised setting}.
We present the results of MPOBERT and other baseline models on GLUE and Squad for fine-tuning in Table~\ref{tab-main_results}. 
Firstly, we evaluate MPOBERT's performance in comparison to other models with similar numbers of parameters. In particular, for tiny models, MPOBERT$_{24}$ outperforms ALBERT$_{24}$, and achieves substantial improvements on both the development set~(85.7 \emph{v.s.} 84.0) and test sets~(82.6 \emph{v.s.} 81.2). This highlights the benefits of increased capacity from layer-specific parameters~(\ie the auxiliary tensors and layer-specific adapters) in MPOBERT. Furthermore, for small and base models, 48-layer MPOBERT consistently achieves better results than T5$_{12}$ and all parameter-efficient models, while also achieving comparable results to other 12-layer PLMs with a reduced number of parameters. This demonstrates the significant benefits of scaling along the model depth with layer-specific parameters in MPOBERT.

Secondly, we assess MPOBERT's parameter efficiency by comparing it to other PLMs within the same model depth. For instance, when considering models with $L$=12 layers, MPOBERT achieves comparable results or even outperforms~(+1.7 for BERT$_{12}$ and +0.4 for XLNet$_{12}$) PLMs while having fewer parameters. This further highlights the advantages of MPOBERT's parameter-efficient approach in constructing deep models.

\paratitle{Multitask Fine-tuning Setting}.
To demonstrate the effectiveness of our proposed parameter-sharing model in learning shared representations across multiple tasks, we fine-tune MPOBERT, BERT and ALBERT on the multitask GLUE benchmark and report the results in Table~\ref{tab:multi-task}. 
Specifically, we design two groups of experiments.~(1) Deep vs. shallow models. Comparing with BERT$_{12}$, MPOBERT$_{48}$ has much deeper Transformer layers but still fewer total parameters~(\ie 75M vs. 110M). We find that MPOBERT$_{48}$ achieves 1.4 points higher on average GLUE score than BERT$_{12}$.~(2) Central tensors sharing vs. all weight sharing. Comparing with ALBERT$_{12}$, MPOBERT$_{12}$ only shares part of weights, \ie central tensors, while ALBERT$_{12}$ shares all of the weights. 
We find that sharing central tensors may effectively improve the average results than sharing all weights~(82.0 \emph{v.s.} 81.4 for MRPC).
\begin{table}[t]
\centering
\small
\begin{tabular}{lrrrr} 
\toprule
    \multicolumn{1}{c}{\multirow{1}{*}{Datasets}} & \multicolumn{1}{c}{B$_{12}$} & \multicolumn{1}{c}{M$_{48}$} & \multicolumn{1}{c}{M$_{12}$} & \multicolumn{1}{c}{A$_{12}$} \\    \midrule
        MNLI~(Acc.)                  &83.9  & 85.4   & 82.8  & 82.7\\
        QNLI~(Acc.)                  &90.8  & 91.1   & 90.0  & 89.4\\
        SST-2~(Acc.)                 &91.7  & 93.0   & 90.9  & 90.6\\
        RTE~(Acc.)                   &81.2  & 82.0   & 79.8  & 79.1\\
        QQP~(Acc.)                   &91.2  & 87.6   & 90.4  & 89.7\\
        CoLA~(Mcc.)                  &53.6  & 54.9   & 45.0  & 35.9\\
        MRPC~(F1)                  &84.2  & 91.8   & 89.9  & 89.2\\
        STS-B~(Spear.)             &87.4  & 89.0   & 86.9  & 87.5\\\midrule
        Avg.                        &83.0  & 84.4   & 82.0  & 80.5\\ 
         {\#To~(M)}                 &110   & 75     & 20    & 11\\
\bottomrule
\end{tabular}
\caption{Performance of multi-task learning on GLUE benchmark obtained by fine-tuning BERT$_{12}$~(B$_{12}$), MPOBERT$_{48}$~(M$_{48}$), MPOBERT$_{12}$~(M$_{12}$) and ALBERT$_{12}$~(A$_{12}$)~(in percent).}
\label{tab:multi-task}
\end{table}

\paratitle{Few-shot Learning Setting}.
\begin{table}[t]
\small
\begin{tabular}{l|ccc|ccc}
\midrule
\multicolumn{1}{c}{}                        & \multicolumn{3}{c}{SST-2} & \multicolumn{3}{c}{MNLI}            \\ \midrule
\multicolumn{1}{c}{Shots~(K)} & 10      & 20    & 30              & 10    & 20    & 30    \\ \midrule
BERT$_{12}$                              & 54.8  & \underline{59.7}  & \underline{61.6}      & \textbf{37.0}  & 35.6  & 35.7  \\ \midrule
ALBERT$_{12}$                            & \underline{56.7}  & 59.3  & 60.0      & 36.3  & 35.6  & \underline{36.5}   \\ \midrule
MPOBERT$_{12}$                           & \textbf{58.9}  & \textbf{65.4}  & \textbf{64.6}      & \underline{36.7}  & \textbf{36.7}  & \textbf{37.1}   \\ \midrule
\end{tabular}
\caption{Comparison of few-shot performance.}
\label{tab-few_shot}
\end{table}
We evaluate the performance of our proposed model, MPOBERT, in few-shot learning setting~\cite{huang-etal-2022-clues} on two tasks, SST-2 and MNLI, using a limited number of labeled examples. Results in Table~\ref{tab-few_shot} show that MPOBERT outperforms BERT, which suffers from over-fitting, and ALBERT, which does not benefit from its reduced number of parameters. These results further demonstrate the superiority of our proposed model in exploiting the potential of large model capacity under limited data scenarios.

\begin{figure}[t]
\centering
\subfigure[Pre-training from scratch]{
\begin{minipage}[t]{0.5\columnwidth}
\label{fig2:left}
\centering
\includegraphics[width=\columnwidth]{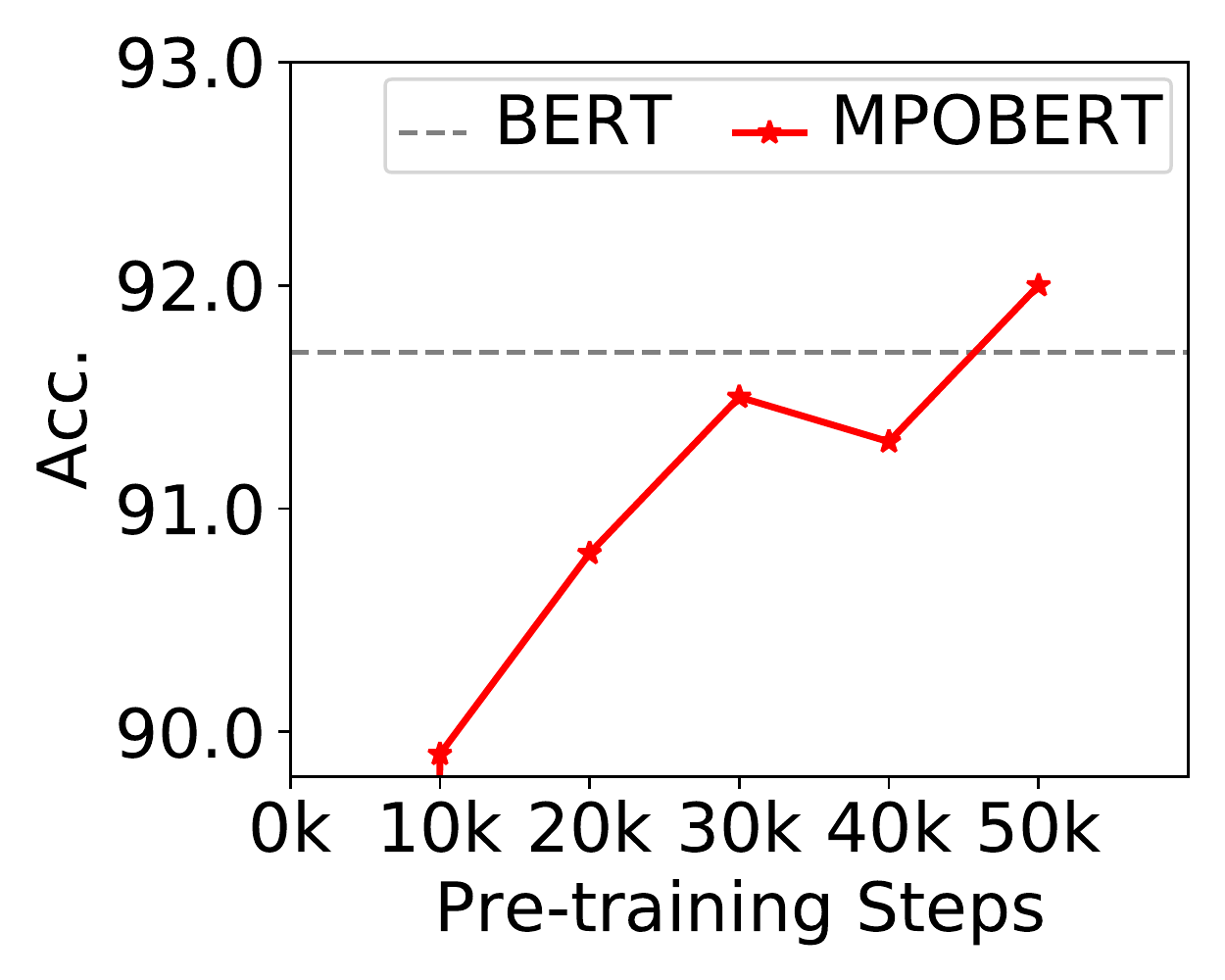} 
\end{minipage}%
}%
\subfigure[Continual Pre-training]{
\begin{minipage}[t]{0.5\columnwidth}
\label{fig2:right}
\centering
\includegraphics[width=\columnwidth]{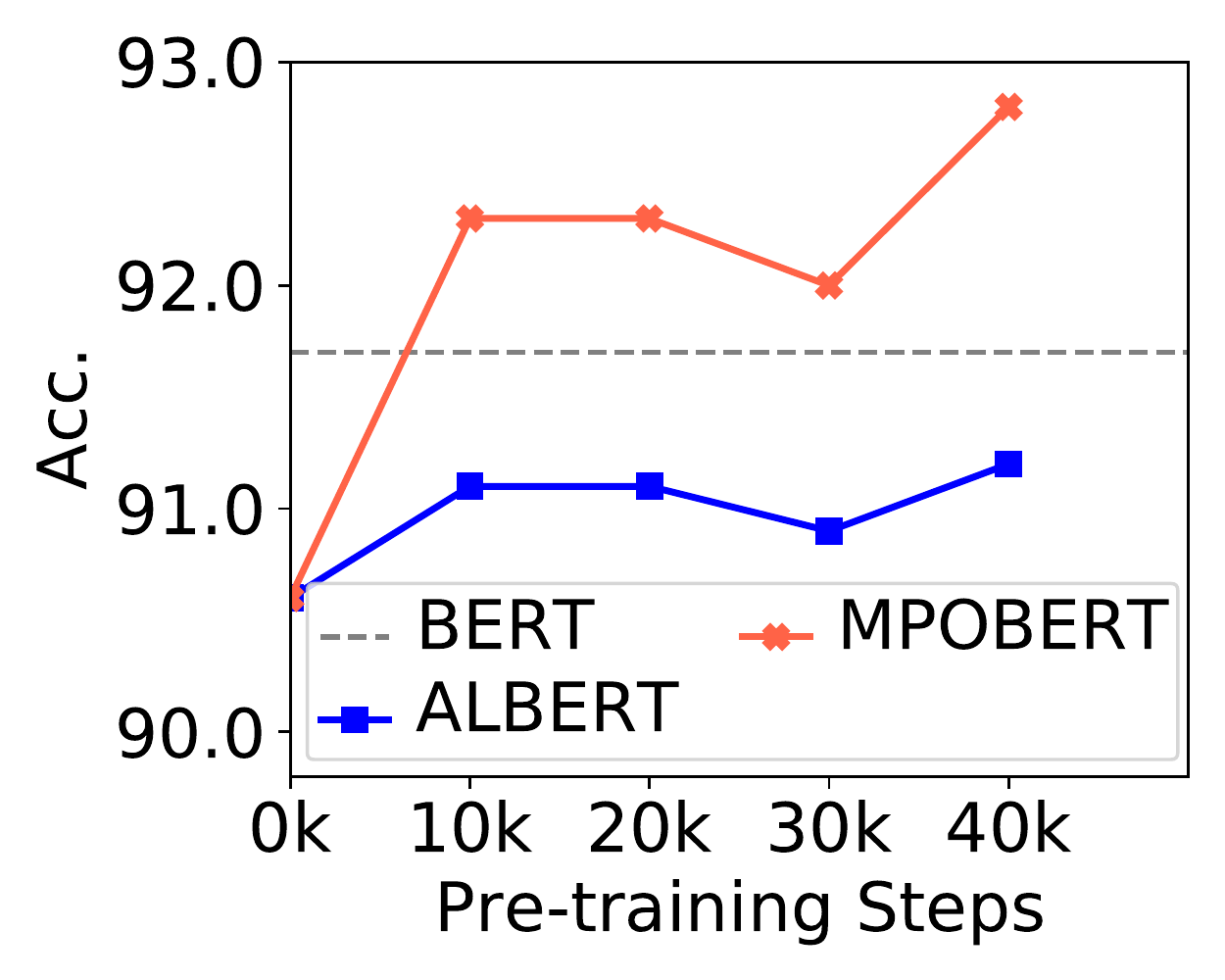} 
\end{minipage}%
}
\caption{Comparison of the SST-2 accuracy achieved through pre-training from scratch and pre-training with the initialization of decomposed ALBERT weights.}
\label{fig:fig2}
\end{figure}
\subsection{Detailed Analysis}
\paratitle{Analysis of Initialization Methods}.
This experiment aims to exclude the effect of initialized pre-trained weights on fine-tuning results. We plot the performance of the model on SST-2 \emph{w.r.t} training steps. In particular, we compare the performance of MPOBERT using different initialization methods (Xavier in Fig.~\ref{fig2:left} and decomposed weights of ALBERT in Fig.~\ref{fig2:right}) for pre-training. The results demonstrate that pre-training MPOBERT from scratch requires around 50$k$ steps to achieve performance comparable to BERT$_{\rm{BASE}}$, while initializing with the decomposed weights of ALBERT significantly accelerates convergence and leads to obvious improvements within the first 10$k$ training steps. In contrast, the gains from continual pre-training for ALBERT are negligible. These results provide assurance that the improvements observed in MPOBERT are not solely attributed to the use of initialized pre-trained weights.
\begin{figure}[htb]
    \centering
    \includegraphics[width=0.48\textwidth]{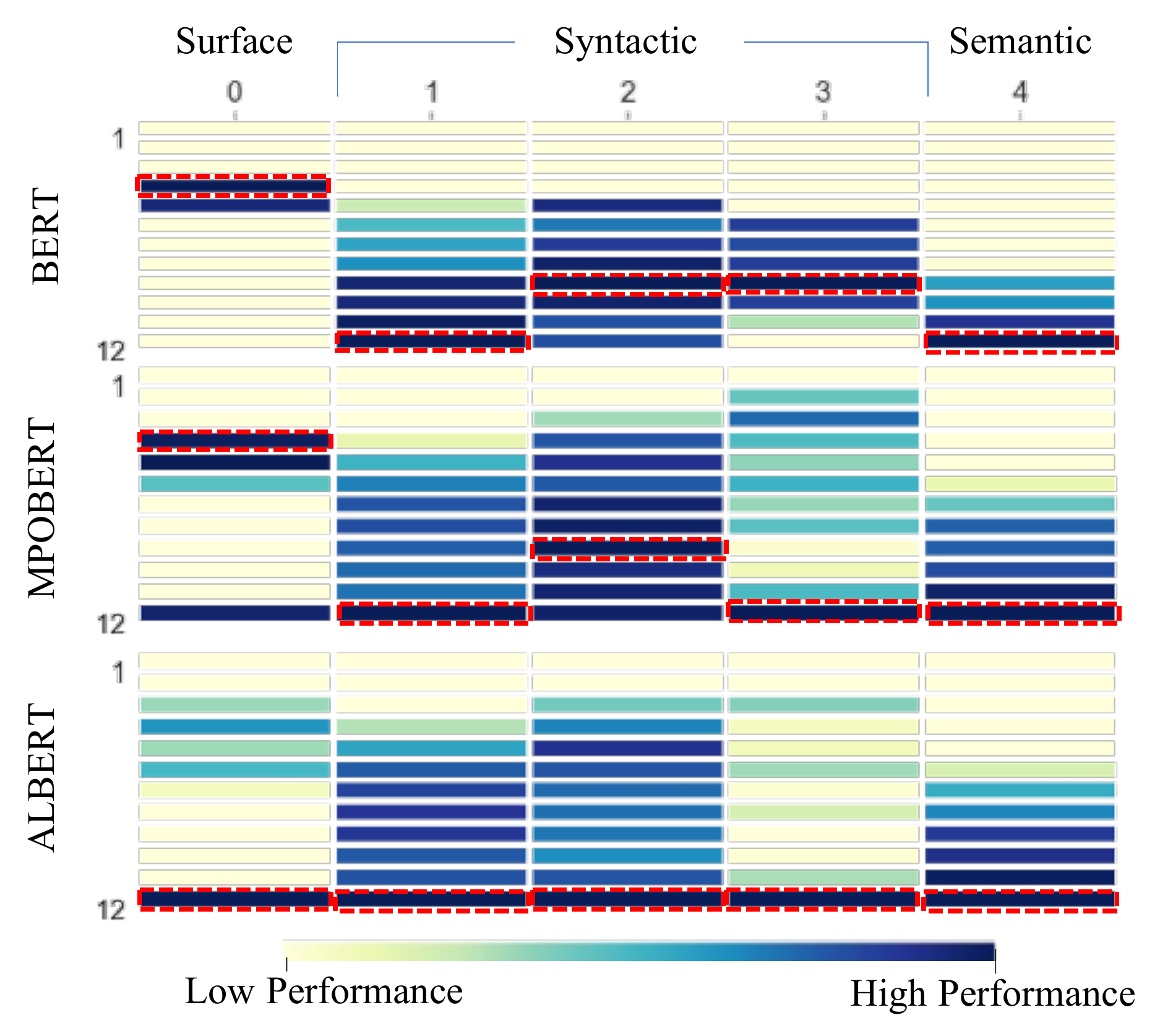}
    \caption{A visualization of layer-wise linguistic patterns. Each column represents a probing task, and each row represents a Transformer layer. The red dashed box indicates the layer that performs best.}
    \label{fig:linguistic}
\end{figure}

\begin{table}[ht]
\centering
\small
\begin{tabular}{lrrrr}                                                                                              
\toprule
Experiment      & SST-2 & RTE     & MRPC  &\#To~(M) \\ \midrule
MPOBERT$_{12}$  & 92.8  & 72.9    & 91.8  & 20.0\\ \midrule
w/o Adapter     & 92.3  & 71.8    & 90.3  & 19.4\\  
w/o PS    & 91.4  & 67.9    & 85.8  & 11.9\\ \bottomrule
\end{tabular}
\caption{Ablation study on the SST-2, RTE, and MRPC datasets~(in percent).}
\label{tab-ablation}
\end{table}
\paratitle{Ablation Analysis}.
To assess the individual impact of the components in our MPOBERT model, we conduct an ablation study by removing either the layer-specific adapter or the cross-layer parameter-sharing strategy. The results, displayed in Table~\ref{tab-ablation}, indicate that the removal of either component results in a decrease in the model's performance, highlighting the importance of both components in our proposed strategy. While the results also indicate that cross-layer parameter sharing plays a more crucial role in the model's performance.
\begin{table}[]
\centering
\small
\begin{tabular}{lrrrrr}
\toprule
Rank                & SST-2     & RTE     & MRPC  &\#To~(M)      \\ \midrule
4                   & 91.9      & 69.7    & 88.2  & 19.7        \\  
8                   & 92.8      & 72.9    & 91.8  & 20.0        \\ 
64                  & 91.6      & 69.3    & 88.1  & 24.3        \\ \bottomrule
\end{tabular}
\caption{Comparison of different adapter ranks on three GLUE tasks~(in percent). ``Rank'' denotes the adapter rank in MPOBERT.}
\label{tab-rank}
\end{table}

\paratitle{Performance Comparison \emph{w.r.t} Adapter Rank}.
To compare the impact of the adapter rank in layer-specific adapters on MPOBERT's performance, we trained MPOBERT with different ranks~(4,8 and 64) and evaluate the model on downstream tasks in Table~\ref{tab-rank}. The results demonstrate that a rank of 8 is sufficient for MPOBERT, which further shows the necessity of layer-specific adapters. However, we also observe a decrease in the performance of the variant with adapter rank 64. This illustrates that further increasing the rank may increase the risk of over-fitting in fine-tuning process. Therefore, we set a rank of 8 for MPOBERT in the main results.

\paratitle{Analysis of Linguistic Patterns}.
To investigate the linguistic patterns captured by MPOBERT, BERT, and ALBERT, we conduct a suite of probing tasks, following the methodology of~\citet{Tinny2019probe}. These tasks are designed to evaluate the encoding of surface, syntactic, and semantic information in the models' representations. The results, shown in Fig.~\ref{fig:linguistic}, reveal that BERT encodes more local syntax in lower layers and more complex semantics in higher layers, while ALBERT does not exhibit such a clear trend.
However, MPOBERT exhibits similar layer-wise behavior to BERT in some tasks~(\ie task 0,2,4), and improved results in lower layers for others~(\ie task 3) which is similar to ALBERT. The result demonstrates that MPOBERT captures linguistic information differently than other models, and its layer-wise parameters play an important role in this difference.
\section{Conclusion}
We develop MPOBERT, a parameter-efficient pre-trained language model that allows for the efficient scaling of deep models without the need for additional parameters or computational resources. 
We achieve this by introducing an MPO-based Transformer layer and sharing the central tensors across layers. During training, we propose initialization methods for the central and auxiliary tensors, which are based on theoretical analysis to address training stability issues. 
The effectiveness of MPOBERT is demonstrated through various experiments, such as supervised, multitasking, and few-shot where it consistently outperforms other competing models.

\section*{Limitations}
The results presented in our study are limited by some natural language processing tasks and datasets that are evaluated, and further research is needed to fully understand the interpretability and robustness of our MPOBERT models. Additionally, there is subjectivity in the selection of downstream tasks and datasets, despite our use of widely recognized categorizations from the literature. 
Furthermore, the computational constraints limited our ability to study the scaling performance of the MPOBERT model at deeper depths such as 96 layers or more. This is an area for future research.
\section*{Ethics Statement}
The use of a large corpus for training large language models may raise ethical concerns, particularly regarding the potential for bias in the data. In our study, we take precautions to minimize this issue by utilizing only standard training data sources, such as BOOKCORPUS and Wikipedia, which are widely used in language model training~\cite{devlin2018bert,lan2019albert}. However, it is important to note that when applying our method to other datasets, the potential bias must be carefully considered and addressed. Further investigation and attention should be given to this issue in future studies.

\bibliography{anthology,custom}
\bibliographystyle{acl_natbib}

\appendix
\clearpage
\section{Appendix}
\label{sec:appendix}
\subsection{Proofs}
\label{app:proof}
\paragraph{Notations.} We denote $\mathcal{L}(\cdot)$ as the loss function. $LN(x)$ as the standard layer normalization with scale $\gamma=1$ and bias $\beta =0$. Let $\mathcal{O}(\cdot)$ denote standard Big-O notation that suppresses multiplicative constants. $\overset{\Theta}{=} $ stands for equal bound of magnitude. 
We aim to study the magnitude of the model updates. We define the model update as $\left\| \bigtriangleup F\right\|$.

\paratitle{Definition}
    $F(x,\theta)$ is updated by $\Theta(\eta)$ per SGD step after initialization as $\eta\to 0$. That is, $\left\| \bigtriangleup F(x)\right\|=\Theta(\eta)$ where $\bigtriangleup F(x)$ can be calculated through $F(x,\theta-\eta\frac{\partial}{\partial \theta}\mathcal{L}(F(x)-y))-F(x;\theta)$.

\begin{theorem}
    Given an $N$-layer transformer-based model $F(x,\theta)(\theta=\{\theta_1, \theta_2, ...,\theta_N\})$, where $\theta_l$ denotes the parameters in $l$-th layer and each sub-layer is normalized with Post-LN: $x_{l+1}=LN(x_l+G_l(x_l,\theta_l))$. In MPOBERT, $\theta_l$ is decomposed by MPO to local tensors: $\theta_l=u_l\cdot c_l\cdot v_l$, and we share $\{c_i\}_{i=1}^{N}$ across $N$ layers: $c_l=c_1, l=1,2,\cdots,N$. Then $\left\| \bigtriangleup F\right\|$ satisfies:
    \begin{align}
    \left\| \bigtriangleup F\right\|\leq
    &\sum_{i=1}^{N}(c_1v_i \left\|u_{i}^*-u_{i}\right\|+ c_1u_i \left\|v_{i}^*-v_{i}\right\| \nonumber\\
    &+ v_iu_i \left\|c_1^*-c_1\right\|)
    \end{align}
    \label{app-thm1}
\end{theorem}
$Proof.$ 
We follow~\cite{zhang2019fixup} and make the following assumptions to simplify the derivations:
\begin{enumerate}
    \item Hidden dimension $d$ equals to $1$;
    \item $var(x+G_l(x))\overset{\Theta}{=}var(x)+var(G_l(x))$;
    \item All relevant weights $\theta$ are positive with magnitude less than $1$.
\end{enumerate}
Given Assumption 1, if $G_l(x)$ is MLP with the weight $\theta_l$, then $G_l(x)\overset{\Theta}{=}\theta_l x$. With assumption 2, we have:
\begin{align}
    x_{l+1}&=f_l(x_l, \theta_l)=\frac{x+G_l(x)}{\sqrt{Var(x+G_l(x))}}\\
    &\overset{\Theta}{=}\frac{1+\theta_l}{\sqrt{1+\theta_l^2}}x_l,
    \label{eq:base}
\end{align}
Then, with Taylor expansion, the model update $\left\|\bigtriangleup F\right\|$satisfies:
\begin{align}
    \left\|\bigtriangleup F\right\|=&\left \|F(x,\theta^*)-F(x, \theta\right \|\nonumber\\
    =&\left\|x_{N+1}^*-x_{N+1}\right\| \nonumber\\
    =&\left\|f(x_{N}^*,\theta_{N}^*) -f(x_{N},\theta_{N})\right\|\nonumber\\
    =&\left \| f(x_{N}^*, U_{N}^*,C_{N}^*,V_{N}^*)\nonumber\right.\\
    &\left.-f(x_N, U_{N},C_{N},V_{N}) \right \| \nonumber\\
    \approx&\left \|\frac{\partial f }{\partial x}(x_{N}^*-x_{N})\nonumber\right.\\
    &\left.+\frac{\partial f}{\partial \theta }\frac{\partial\theta}{\partial U_{N}}(U_{N}^*-U_{N})^T \nonumber\right.\\
    &\left.+\frac{\partial f}{\partial \theta }\frac{\partial\theta}{\partial C_{N}}(C_{N}^*-C_{N})^T\nonumber\right.\\
    &\left.+\frac{\partial f}{\partial \theta }\frac{\partial\theta}{\partial V_{N}}(V_{N}^*-V_{N})^T  \right \|
    \label{eq:9}
\end{align}
With Eq.~\eqref{eq:base}, the magnitude of $\frac{\partial f_l}{\partial x}$ and $\frac{\partial f_l}{\partial \theta}$ is bounded by:
\begin{align}
    & \frac{\partial f_l}{\partial x}\overset{\Theta}{=}\frac{1+\theta_l}{\sqrt{1+\theta_l^2}} \\
    & \frac{\partial f_l}{\partial \theta_l}\overset{\Theta}{=}\frac{1-\theta_l}{(1+\theta_l^2)^{\frac{3}{2}}}x_l
\end{align}
Since we apply MPO decomposition to $\theta_l$, we get:
\begin{align}
    \theta_l=U_l\cdot C_l \cdot V_l
\end{align}
For simplicity, we reduce the matrices $U$,$C$,$V$ to the scalars $u$,$c$,$v$. 
Thus with Assumption 3, Eq.~\eqref{eq:9} is reformulated as:
Finally, with Assumption 3 we have:
\begin{align}
    \left\|\bigtriangleup F \right\|=
    &\left\| x_{N+1}^*-x_{N+1} \right\| \\
    \leq &\sum_{i=1}^{N}\frac{1-u_ic_1v_i}{({{1+u_{i}^2c_1^2v_{i}^2}})^{\frac{3}{2}}}(c_1v_i \left\|u_{i}^*-u_{i}\right\|\nonumber\\
    &+ c_1u_i \left\|v_{i}^*-v_{i}\right\|) + v_iu_i \left\|c_1^*-c_1\right\|) \nonumber\\
    \approx&\sum_{i=1}^{N}(c_1v_i \left\|u_{i}^*-u_{i}\right\|+ c_1u_i \left\|v_{i}^*-v_{i}\right\| \nonumber\\
    &+ v_iu_i \left\|c_1^*-c_1\right\|)
\end{align}
\rightline{$\Box$}
\begin{corollary}
\label{app-thm2}
    Given that we initialise $c_1$ in MPOBERT with well-trained weights, it is reasonable to assume that updates of $c_1$ are well-bounded.
    Then $\bigtriangleup F$ satisfies $\left\| \bigtriangleup F\right\|=\mathcal{O}(1)$ when for all $i=1,\cdots,N$:
    \begin{equation}
        (v_i^2+u_i^2)(u_Nv_N)=\mathcal{O}(\frac{1}{N})
    \end{equation}
\end{corollary}

$Proof.$ 
For an $N$-layer MPOBERT, we have:
\begin{align}
    \left \|\bigtriangleup F\right \|
    \leq &\sum_{i=1}^{N}(v_i \left\|u_{i}^*-u_{i}\right\|+{u_i \left\|v_{i}^*-v_{i}\right\|}) \\
    \leq &\eta\sum_{i=1}^{N}(v_i \left\|\frac{\partial \mathcal{L}}{\partial F}\right\|\cdot \left\|\frac{\partial F}{\partial \theta_i}\right\|\cdot \left\|\frac{\partial \theta_i}{\partial u_i}\right\|\nonumber\\
    &+{u_i \left\|\frac{\partial \mathcal{L}}{\partial F}\right\|\cdot \left\|\frac{\partial F}{\partial \theta_i}\right\|\cdot\left\|\frac{\partial \theta_i}{\partial v_i}\right\|})
\end{align}
By assumption $\left\| \frac{\partial \mathcal{L}}{\partial F}\right\|=\mathcal{O}(1)$ and $\left\|\frac{\partial F}{\partial {\theta_i}} \right\|\leq\left\| \frac{\partial F}{\partial \theta_N} \right\|\overset{\Theta}{=}\left\| \theta_{N}\right\|$, we achieve:
\begin{align}
    &\eta\sum_{i=1}^{N}(v_i \left\|\frac{\partial \mathcal{L}}{\partial F}\right\|\cdot \left\|\frac{\partial F}{\partial \theta_i}\right\|\cdot \left\|\frac{\partial \theta_i}{\partial u_i}\right\|\nonumber\\
    &+{u_i \left\|\frac{\partial \mathcal{L}}{\partial F}\right\|\cdot \left\|\frac{\partial F}{\partial \theta_i}\right\|\cdot\left\|\frac{\partial \theta_i}{\partial v_i}\right\|})\\
    =&\eta\sum_{i=1}^{N}(v_i^2u_Nv_N+u_i^2u_Nv_N) \nonumber\\
    = &\mathcal{O}(\sum_{i=1}^{N}(v_i^2+u_i^2)(u_Nv_N))=\mathcal{O}(1),
\end{align} 
\label{eq:bound}
Finally, we achieve:
\begin{equation}
    (v_i^2+u_i^2)(u_Nv_N)=\mathcal{O}(\frac{1}{N})
\end{equation}

Due to symmetry, we set $u_i=u$, $v_i=v$. Thus, from~\ref{eq:bound}, we set $u=v=(2N)^{-\frac{1}{4}}$ to achieve to bound the magnitudes of each update to be independent of model depth $N$, \ie $\left\| \bigtriangleup F\right\|=\mathcal{O}(1)$.
\rightline{$\Box$}

\begin{algorithm}[htb]
    \caption{The MPOBERT training procedure.}
    \begin{algorithmic}[1] 
    \small
        \Require $\Matrix{W}^{(l)}$: Weight matrix of $l$-th layer in MPOBERT.
        $\Matrix{W}_{A}^{(0)}$: Pre-trained weight matrix in ALBERT.
        $\Matrix{U}^{(l)}$ and $\Matrix{D}^{(l)}$: Matrices in low-rank adapter.
        $\eta$: Learning rate.
        $\mathcal{L}$: Stochastic objection function.
        $L$: Model layers number.
        \Statex (MPO decomposition)
        \State
        $\{\Tensor{A}_1^{(l)},\Tensor{A}_2^{(l)},\Tensor{C}^{(l)},\Tensor{A}_3^{(l)},\Tensor{A}_4^{(l)}\}$ $\gets$ MPO ($\Matrix{W}^{(l)}$)
        \State
        $\{\Tensor{A}_1^{(0)},\Tensor{A}_2^{(0)},\Tensor{C}^{(0)},\Tensor{A}_3^{(0)},\Tensor{A}_4^{(0)}\}$ $\gets$ MPO ($\Matrix{W}_{A}^{(0)}$)
        \Statex (Initialization Procedure)
        \For {$0<l\leq 24$}
            \State  $\Tensor{C}^{(l)} \gets \Tensor{C}^{(0)} , 
             \{\Tensor{A}_{j}^{(l)}\}_{j=1}^{4} \gets \{\Tensor{A}_{j}^{(0)}\}_{j=1}^{4}$ 
        \EndFor
        \For {$ 24<l\leq L$}
            \State  $\Tensor{C}^{(l)} \gets \Tensor{C}^{(0)} ,    \{\Tensor{A}_{j}^{(l)}\}_{j=1}^{4} \gets \{(2L)^{-\frac{1}{4}}\Tensor{A}_{j}^{(0)}\}_{j=1}^{4}$ 
        \EndFor
        \State $\Matrix{U}^{(l)} \gets \Matrix{0}$, $\Matrix{D}^{(l)} \gets \mathcal{N}(0, \sigma^2)$
        \State $\Matrix{W}^{(l)}=\Tensor{A}_1^{(l)}\Tensor{A}_2^{(l)}\Tensor{C}^{(l)}\Tensor{A}_3^{(l)}\Tensor{A}_4^{(l)}+\Matrix{W}_{Adapter}^{(l)}$
        \Statex (Training procedure with mixed precision and fused implementation techniques.)
        \While {not converged}
            \State $t \gets t+1$
            \State $g_t \gets \frac{\partial\mathcal{L}(\Matrix{W}^{(l)}_t)}{\partial(\Matrix{W}^{(l)}_t)}$
            \State $\Matrix{W}^{(l)}_t \gets \Matrix{W}^{(l)}_{t-1} - \eta \cdot g_t$
        \EndWhile
        \State \Return Converged model
    \end{algorithmic}
\label{alg-overall-process}
\end{algorithm}
\subsection{Training Details}
\label{add-trianing-detail}

\subsubsection{Details of Training}
Here we describe the details of the pre-training process in Algorithm~\ref{alg-overall-process}. 
For pre-training, we tune the learning rate in the range of [$1.0\times 10^{-5}$, $1.0\times 10^{-6}$] and use the LAMB optimizer~\cite{you2020lamb}. Since fine-tuning is typically fast, we run an exhaustive parameter search~(\ie learning rate in the range of [$2.0\times 10^{-4}$, $2.0\times 10^{-6}$], batch size in \{8,16,32\}) and choose the model that performs best on the development set to make predictions on the test set.

\subsubsection{Details of Training Configurations}
In this part, we list the training configurations of MPOBERT and other representative PLMs in Table~\ref{tab-strongest_variants}.
\begin{table*}[t]
\centering
\begin{tabular}{lrrrcrrr}                       
\toprule
Models    & \#To~(M) & Depth & Samples  & Training time  & GLUR Dev.  &GLUE Test      \\ \midrule
T5$\rm{_{11B}}$        & 11000  & 24     & -   & -    & - & 89.0       \\  \\
T5$\rm{_{BASE}}$        & 220  & 24     & 128$\times$ 524$k$   & \multirow{2}{*}{\thead{16 TPU v3\\ 1 Day~(t5-base)}}  & 84.1    & 82.5       \\  \\
BERT$\rm{_{LARGE}}$      & 330     & 24     & 256$\times$ 1000$k$   & \multirow{2}{*}{\thead{16 Cloud TPUs\\ 4 Days}}  & 84.1    & 81.6       \\  \\
ALBERT$\rm{_{XXLARGE}}$    & 235     & 1     & 4096$\times$ 1.5$M$   & \multirow{2}{*}{\thead{TPU v3\\ 16 Days}}  & 90.0    & -       \\  \\
BART$\rm{_{LARGE}}$      & 407     & 24     & 8000$\times$ 500$k$   & -  & 88.8    & -       \\  \\
RoBERTa$\rm{_{LARGE}}$   & 355     & 24     & 8000$\times$ 500$k$   & \multirow{2}{*}{\thead{1024 V100 GPUs\\ 1 Day}}  & 88.9    & -       \\  \\
XLNet$\rm{_{LARGE}}$     & 361     & 24     & 8192$\times$ 500$k$   & \multirow{2}{*}{\thead{512 TPU v3\\ 5.5 Days}}  & 87.4    & -       \\  \\
MPOBERT$_{48+}$   & 102     & 48    & 4096$\times$ 10$k$    & \multirow{2}{*}{\thead{8 V100 GPUs\\ 3.8 Days}}    & 85.6  & 81.7  \\ \\ \bottomrule
\end{tabular}
\caption{Comparison with the strongest variants of popular PLMs. Since T5$\rm{_{11B}}$ has far more parameters than other candidates, it's reasonable to use T5$\rm{_{base}}$ for a fair comparison.}
\label{tab-strongest_variants}
\end{table*}

\subsection{Experimental Details}
\subsubsection{Details of Fine-tuning Datasets}
\label{add-detail_dataset}
GLUE benchmark covers multiple datasets~(MNLI, QNLI, QQP, CoLA, RTE, MRPC, SST-2)~\footnote{In line with~\citet{raffel2020exploring}, we do not test WNLI due to its adversarial character with respect to the training set.}. 
The SQuAD is a collection of 100$k$ crowd-sourced question/answer pairs. Given a question and a passage, the task is to predict the answer text span in the passage. 

\subsubsection{Details of Evaluation Metrics}
\label{add-detail_metric}
Following~\citet{gao2022parameter}, we employ Matthew's correlation for CoLA, Spearman for STS-B, F1 for MRPC, and accuracy for the remaining tasks as the metrics for the GLUE benchmark.
We compute and present the average scores across all test samples for each of the aforementioned metrics.

\subsubsection{Details of Baseline Models}
\label{add-detail_baseline}
We compare our proposed MPOBERT to the existing competitive deep PLMs and parameter-efficient models. In order to make fair comparisons, we divide the models into three major categories based on their model sizes: 
$\bullet$~{Tiny Models~(\rm{\#To < 50M}).} ALBERT$_{12}$~\cite{lan2019albert} is the most representative PLM that achieves competitive results with only 11M.

$\bullet$~{Small models~(50M< \#To <100M).} T5$_{12}$ is a small variant of T5~\cite{raffel2020exploring} which has only 6 encoder layers and 6 decoder layers. In addition, there are three parameter-efficient Transformer models that have similar parameters, namely MobileBERT~\citep{sun2020mobilebert}, DistilBERT~\citep{sanh2019distilbert} and TinyBERT~\citep{jiao2019tinybert}. We compare with these compressed models to show the benefit of scaling to deeper models over compressing large models to small variants.

$\bullet$~{Base models~(\#To > 100M).} We compare with BERT$_{12}$, XLNet$_{12}$, RoBERTa$_{12}$ and BART$_{12}$ for this category. Note that we only include the base variants that have similar model sizes in order to make a fair comparison. More details about the comparison with the strongest variants are described in Appendix~\ref{app-exp}.

\label{app-exp}

\end{document}